\documentclass{article}

\usepackage{arxiv}

\usepackage[utf8]{inputenc} 
\usepackage[T1]{fontenc}    
\usepackage{hyperref}       
\usepackage{url}            
\usepackage{booktabs}       
\usepackage{amsfonts}       
\usepackage{nicefrac}       
\usepackage{microtype}      
\usepackage{amsmath} 
\usepackage{cleveref}       
\usepackage{lipsum}         
\usepackage{graphicx}
\usepackage{natbib}
\usepackage{doi}
\usepackage{enumitem}

\title{Penrose Tiled Low-Rank Compression and Section-Wise Q\&A Fine-Tuning: 
A General Framework for Domain-Specific Large Language Model Adaptation}

\author{
Chuan-Wei Kuo$^{1}$\thanks{Equal contribution.}, Siyu Chen$^{1,2}$\footnotemark[1], Chenqi Yan$^{3}$, and Yu Yang Fredrik Liu$^{1}$\thanks{Corresponding author.} \\
$^{1}$DeepVerse, Shanghai, China \\
$^{2}$TCM Group, Cavendish Laboratory, University of Cambridge, Cambridge, United Kingdom \\
$^{3}$School of Electronic Information and Electrical Engineering, Shanghai Jiao Tong University, Shanghai, China \\
\texttt{\{auberonkuo, siyu, yuyang\}@deepverse.tech}, \texttt{yanchenqi@sjtu.edu.cn}
}

\date{}

\renewcommand{\headeright}{DeepVerse}
\renewcommand{\undertitle}{Preprint Version}
\renewcommand{\shorttitle}{A General Framework for Domain-Specific Large Language Model Adaptation}

\hypersetup{
  pdftitle={Penrose-Like Low-Rank Tiling and Spectral Compression: A Visionary Framework for Open Exploration in Large Language Model Adaptation},
  pdfsubject={cs.LG, cs.AI},
  pdfauthor={Chuan-Wei Kuo, Siyu Chen, Yu Yang Fredrik Liu},
  pdfkeywords={Large Language Models, Low-Rank Decomposition, Penrose Tiling, Frequency-Domain Compression, Domain Adaptation},
}

\begin{document}
\maketitle

\begin{abstract}
Large language models (LLMs) hold great promise for specialized scientific domains such as materials science, yet adapting them efficiently and accurately to domain-specific knowledge remains challenging due to limited data and high knowledge density. We propose a two-stage framework that combines structured model compression with a scientific fine-tuning regimen to address this challenge. In the compression stage, we decompose the LLM’s weight matrices into local low-rank "rank blocks" and arrange these blocks in a Penrose-like non-periodic tiling pattern. Each block is then compacted via spectral transformations (e.g., discrete cosine or Fourier transforms), and a Kullback–Leibler (KL) divergence-based alignment loss preserves the distributional similarity between the compressed model’s representations and those of the original full model. In the adaptation stage, the compressed model is further tuned using a human-like scientific reading protocol: it processes technical materials science documents section by section, engaging in a structured question-and-answer routine for each section. This section-wise Q\&A fine-tuning strategy extracts explicit reasoning traces and gradually injects domain knowledge, while minimizing catastrophic forgetting of the model’s general language capabilities. By balancing efficient compression with targeted adaptation, our two-stage approach enables precise specialization of LLMs to high-value domains under data-scarce conditions. We present this principled yet exploratory pipeline and outline its potential for advancing materials science knowledge integration, laying the groundwork for comprehensive empirical evaluation in future work.
\end{abstract}



\section{Introduction}

Large Language Models (LLMs) have showcased unprecedented capabilities in a wide range of natural language processing tasks, with parameter counts often reaching hundreds of billions. However, scaling to such massive models comes at steep computational costs and substantial storage demands. Furthermore, it remains challenging to adapt these models to specialized domains such as materials science, where data are relatively scarce but knowledge density is high. Traditional compression techniques (e.g., pruning, quantization, or knowledge distillation) and parameter-efficient tuning methods (e.g., LoRA \citep{hu2022lora}) have partly addressed this issue, yet they typically enforce uniform or block-circulant assumptions, risking the loss of intricate patterns embedded in large Transformer-based architectures.

\paragraph{Penrose Tiled Low-Rank Compression.}
In this paper, we propose a two-stage framework that systematically compresses and then domain-specializes an LLM. The first stage comprises a \emph{Penrose Tiled Low-Rank Compression} procedure. We treat the model’s massive weight matrices as a ``giant puzzle'' and factor them into \emph{rank blocks}---submatrices of lower (or partial-high) rank that capture essential directions of variability. Unlike standard block partitioning, these rank blocks are arranged in a \textbf{Penrose tiling} (or quasicrystal-like) fashion, introducing non-periodic yet locally consistent structure across the parameter space. Inspired by quasicrystals in physics, the Penrose layout endows our compression scheme with flexibility to capture local correlations without enforcing a repetitive global grid.

We further \textbf{compress each rank block in the frequency domain} via transforms such as the Discrete Cosine Transform (DCT) or fast Fourier Transform (FFT). This step prunes high-frequency coefficients and preserves the essential lower-frequency components. To ensure the compressed model retains semantic fidelity, we incorporate a \textbf{KL-divergence alignment loss} that compares hidden-layer distributions of the compressed model with those of the original full model. Intuitively, aligning these distributional signatures helps the compressed model maintain key linguistic and reasoning capabilities, despite discarding many parameters. While the majority of submatrices remain low-rank, we allow certain ``high-rank'' blocks in regions where preserving intricate weight structures proves critical, thus offering a dynamic balance between compression and performance retention.

\paragraph{Section-Wise Q\&A Fine-Tuning.}
After compressing the model, we then perform \textbf{section-wise Q\&A fine-tuning} tailored to domain knowledge injection. In contrast to conventional one-pass fine-tuning, our approach treats domain documents---e.g., materials science papers---as a sequence of sections or paragraphs, prompting the model with relevant questions at each step. For every chunk of text, the model generates or answers domain-specific queries, optionally accompanied by reasoning traces \citep{wei2022chain}. This \emph{human-like reading} regimen serves multiple purposes:
\begin{itemize}
    \item \textbf{Progressive Knowledge Integration:} By iterating through sections, the model incrementally absorbs specialized terminology and concepts, mirroring the way a human expert studies a scientific text.
    \item \textbf{Reduced Catastrophic Forgetting:} Frequent Q\&A cycles revisit both domain-specific ideas and the model’s general language skills, preventing the overshadowing of broad knowledge by narrow details.
    \item \textbf{Enhanced Interpretability:} Logging intermediate chain-of-thought aids in diagnosing how the compressed model consolidates new material, crucial for tasks where correctness and traceability are paramount.
\end{itemize}

\paragraph{Materials Science as a Use Case.}
We demonstrate this two-stage framework---Penrose tiled compression plus Q\&A fine-tuning---in the context of materials science, a domain where data are comparatively limited yet highly specialized \citep{tshitoyan2019unsupervised, gupta2022matscibert}. This field presents an excellent stress test for domain adaptation: researchers must parse intricate texts involving crystal lattices, phase diagrams, and chemical processes, all under computational constraints. By compressing an LLM via rank blocks and frequency transforms while preserving distributional integrity, we obtain a lightweight yet robust model capable of tackling section-based reading and question answering. This method offers a blueprint for broader scientific domains (e.g., bioinformatics or theoretical physics), where data scarcity, knowledge specificity, and interpretability demand innovative compression and adaptation strategies.

\paragraph{Key Contributions.}
Overall, the contributions of this paper include:
\begin{enumerate}
    \item \textbf{Penrose Tiled Low-Rank Decomposition:} A novel non-periodic partitioning of large weight matrices into rank blocks, inspired by quasicrystalline geometry, aiming to reconcile local structure with global flexibility.
    \item \textbf{Frequency-Domain Compression \& Distribution Alignment:} Applying transforms (e.g., DCT) to prune high-frequency coefficients, combined with a KL-divergence-based loss to preserve the original model’s hidden representations.
    \item \textbf{Section-Wise Q\&A Fine-Tuning:} A reading-centric adaptation strategy for materials science literature, reducing catastrophic forgetting and offering clearer interpretability through explicit reasoning traces.
    \item \textbf{Practical Case Study in Materials Science:} We illustrate how to adapt and evaluate an LLM in a high-value, data-scarce scientific domain, laying groundwork for similar approaches in other specialized fields.
\end{enumerate}

In the following sections, we first discuss relevant prior work on low-rank model compression, quasicrystal-inspired tiling, and domain-specific LLM adaptation (\S\ref{sec:related_work}). We then detail our compression methodology and frequency-domain alignment (\S\ref{sec:method_compression}), followed by the Q\&A fine-tuning procedure (\S\ref{sec:method_finetuning}). Next, we present our experimental setup and an initial exploration in materials science (\S\ref{sec:experiments}), integrate broader limitations and future directions into an in-depth discussion (\S\ref{sec:discussion}), and finally conclude with overall implications for large-model specialization in \S\ref{sec:conclusion}.

\section{Related Work}
\label{sec:related_work}

Our proposed framework intersects multiple research fronts, including low-rank or hybrid-rank factorization for model compression, distribution alignment with KL-divergence, quasicrystal-inspired tiling, frequency-domain transforms, and domain adaptation through human-like reading. Below, we summarize the most relevant developments in each area.

\subsection{Low-Rank and Hybrid-Rank Decomposition for LLM Compression}
\label{sec:related_lowrank}

The notion of compressing large language models (LLMs) using low-rank factorizations stems from the observation that many neural network weight matrices exhibit redundant or correlated dimensions \citep{denil2013predicting, jaderberg2014speeding}. Early works applied singular value decomposition (SVD) to prune smaller singular values, thereby reducing parameter counts \citep{denton2014exploiting}. More recently, \textbf{LoRA} (Low-Rank Adaptation) introduced a parameter-efficient tuning paradigm for LLMs by reparameterizing certain weight updates as products of small rank-$r$ matrices \citep{hu2022lora}. This approach drastically cuts down the number of trainable parameters while preserving performance in downstream tasks.

However, real-world Transformers often have complex, partially high-rank weight structures that pure low-rank models fail to capture. To address this, hybrid approaches combine low-rank submatrices with residuals or sparse structures. For example, \textbf{LoSparse} decomposes each weight matrix into a low-rank plus a sparse component to better accommodate high-rank or unstructured signals \citep{li2023losparse}. Similarly, \textbf{CALDERA} represents each matrix as a combination of a quantized residual term $\mathbf{Q}$ and two low-rank factors $\mathbf{L}\mathbf{R}$, achieving better trade-offs under memory constraints \citep{saha2024compressing}. Meanwhile, \textbf{MoDeGPT} partitions large Transformer blocks into finer modules and applies factorization or Nyström-based approximations within each module, compressing up to 25--30\% of parameters while retaining strong zero-shot performance \citep{lin2024modegpt}.

Despite these advances, existing methods generally rely on layer-by-layer or uniform block structures. Our work diverges by treating each low-rank (or partially high-rank) \emph{rank block} as a puzzle piece, potentially rearranged via a quasicrystal-inspired tiling. We posit that a \emph{non-periodic} organization may capture emergent, fine-grained correlations that conventional schemes overlook.

\subsection{Distribution Alignment via KL-Divergence and Related Metrics}
\label{sec:related_kl}

Maintaining the original model’s representation distribution is often essential for avoiding catastrophic performance drops after compression. A well-known strategy in \emph{knowledge distillation} is to use a Kullback–Leibler (KL) divergence loss to align the logits or intermediate-layer distributions between teacher and student models \citep{hinton2015distilling}. Several studies have extended this idea beyond logits to attention maps \citep{zagoruyko2016paying} or hidden states \citep{romero2014fitnets}.

In the context of LLM compression, \citet{kim2021bert} used a KL-based objective to preserve token-level representations, while \citet{mukherjee2023distilling} leveraged cross-entropy or Jensen–Shannon divergence to retain multi-head attention patterns. Our approach similarly incorporates a \textbf{KL-divergence alignment loss}, but it targets the discrepancy between hidden-layer activations of the \emph{compressed} (Penrose-tiled) model and those of the \emph{original} full model. By aligning distributions at an internal level, we aim to retain subtle linguistic and reasoning capabilities despite heavily restructuring the parameters. Although KL is our primary focus, other statistical distances (e.g., Wasserstein, Jensen–Shannon) could serve a similar role of enforcing distributional fidelity.

\subsection{Penrose Tiling and Non-Periodic Factorization}
\label{sec:related_penrose}

While structured factorization methods (e.g., block-circulant or Toeplitz matrices) have been explored for efficient neural network design \citep{ding2017c, sindhwani2015structured}, these approaches typically assume globally repetitive patterns. In contrast, a \textbf{Penrose tiling} uses a finite set of tile shapes arranged in a \emph{non-periodic} yet locally consistent manner. The concept originates from quasicrystals in physics, where atoms exhibit short-range regularity without repeating over large scales \citep{levine1984quasicrystals}.

To our knowledge, no prior work explicitly models LLM weight matrices as a Penrose- or quasicrystal-like layout. However, analogies to fractal or aperiodic structures have occasionally arisen in discussions of deep network connectivity \citep{larsson2016fractalnet}. We propose extending this perspective to the parameter space itself, hypothesizing that large Transformers may exhibit “quasicrystalline” sub-block patterns across attention heads or feed-forward layers. Should an aperiodic factorization prove advantageous, it might reduce redundancy while preserving essential local correlations more flexibly than strict periodic grids. Empirically validating these claims remains an open challenge.

\subsection{Frequency-Domain Compression for Neural Weights}
\label{sec:related_freq}

Frequency-domain transforms (e.g., Fourier, Cosine, Wavelet) have been extensively applied to audio/image compression, exploiting the fact that many signals are concentrated in lower frequencies. In neural networks, \citet{chen2015compressing} proposed \textbf{HashedNets} and \textbf{FreshNets}, which group and quantize CNN filter coefficients in the frequency domain. \citet{laude2018adaptive} extended such ideas to fully connected layers, employing DCT-based transform coding to prune insignificant coefficients.

Although these methods demonstrate that “\emph{many weights are effectively high-frequency noise},” most frequency-domain compression research still targets convolutional architectures or smaller-scale networks. Adapting it to LLMs raises non-trivial questions: How do we vectorize multi-head attention matrices or transformer feed-forward blocks for spectral analysis? Which frequency bands are most critical for retaining semantic integrity? Our framework’s \emph{rank-block + spectral transform} step is a step toward bridging this gap. By first factoring out principal directions (low-rank) and then pruned in the frequency domain, we hope to reveal new possibilities for compressing extremely large weight matrices.

\subsection{Domain Adaptation and Human-Like Reading Protocols}
\label{sec:related_adaptation}

Finally, the second stage of our framework aligns with a broader trend in \textbf{domain adaptation} for LLMs. Existing approaches include further pre-training on domain corpora \citep{gururangan2020don}, lightweight prompt tuning \citep{lester2021power}, adapter modules \citep{pfeiffer2020adapterhub}, or LoRA-based incremental updates \citep{hu2022lora}. Yet these often treat the domain corpus as a monolithic block of text, ignoring potential hierarchical structure in scientific or technical writing.

In contrast, \citet{wei2022chain} highlight the benefits of generating intermediate reasoning steps (chain-of-thought), which can improve factual correctness and interpretability. Likewise, some reading comprehension research organizes content into question-answer pairs to mimic how humans study or summarize documents \citep{kwiatkowski2019natural}. Building on these insights, our \textbf{section-wise Q\&A fine-tuning} trains an LLM to parse specialized texts in a \emph{human-like} manner---section by section, question by question. This iterative reading protocol can reduce catastrophic forgetting by repetitively anchoring new domain knowledge to previously learned background, potentially leading to more stable and interpretable domain adaptation.

\subsection*{Summary}
Taken together, our work extends classical low-rank compression into an aperiodic, rank-block perspective (\S\ref{sec:related_lowrank}), reinforced by distribution alignment via KL-divergence (\S\ref{sec:related_kl}) and frequency-domain techniques (\S\ref{sec:related_freq}). By drawing inspiration from Penrose tiling (\S\ref{sec:related_penrose}), we aim to explore whether a non-periodic arrangement of blocks captures complex weight correlations more naturally than conventional uniform partitions. We then combine this structured compression with a reading-centric domain adaptation paradigm (\S\ref{sec:related_adaptation}) that treats scientific documents as iterative Q\&A sessions. In the following sections, we describe our methodology in detail and present a case study in materials science, illustrating how these disparate ideas may coalesce into a coherent, next-generation framework for efficient and interpretable LLM specialization.

\section{Methodology}
\label{sec:methodology}

Our proposed methodology builds on the key ideas outlined in Sections~\ref{sec:related_lowrank}--\ref{sec:related_adaptation}, transforming them into a two-stage pipeline for compressing and adapting large language models (LLMs). The major components of our framework are:
\begin{enumerate}
    \item \textbf{Factorizing} the pretrained LLM’s weight matrices into local \emph{rank blocks};
    \item \textbf{Arranging} these blocks non-periodically via a Penrose tiling;
    \item \textbf{Performing} frequency-domain compression with distribution alignment;
    \item \textbf{Applying} a section-wise Q\&A regimen to inject specialized domain knowledge.
\end{enumerate}

In the subsections that follow, we detail each step, highlighting the mathematical rationale, practical considerations, and open challenges. While we primarily illustrate the pipeline with standard transformer layers, the approach can be adapted to other architectures (e.g., convolution-based backbones or hybrid designs).

\subsection{Stage I: Penrose-Tiled Low-Rank Compression}
\label{sec:method_compression}

\subsubsection{Rank Block Decomposition}
\label{sec:rank_decomposition}
Consider a pretrained LLM with $L$ transformer layers. Each layer typically contains multiple large weight matrices, such as the query/key/value projections in multi-head attention or the feed-forward sublayers (e.g., $\mathbf{W}_{\mathrm{in}}, \mathbf{W}_{\mathrm{out}} \in \mathbb{R}^{m \times n}$). To reduce redundancy, we approximate each matrix $\mathbf{W}$ via low-rank factorization:
\begin{equation}
    \mathbf{W} \approx \mathbf{U} \mathbf{V}^\top,
    \quad \mathbf{U} \in \mathbb{R}^{m \times r}, \;\mathbf{V} \in \mathbb{R}^{n \times r},
\end{equation}
where $r \ll \min(m,n)$. Instead of storing $(\mathbf{U}, \mathbf{V})$ in a monolithic form, we \emph{partition} these factors into $K$ smaller submatrices:
\begin{align}
    \mathbf{U} &= \big[\mathbf{U}_1,\,\mathbf{U}_2,\,\dots,\mathbf{U}_K\big], 
    \quad
    \mathbf{V} = \big[\mathbf{V}_1,\,\mathbf{V}_2,\,\dots,\mathbf{V}_K\big],
    \\
    \mathbf{B}_i &\approx \mathbf{U}_i \,\mathbf{V}_i^\top, \quad i=1,\ldots,K.
\end{align}
Each $\mathbf{B}_i$, which we refer to as a \emph{rank block}, can be viewed as a local puzzle piece in the global weight matrix. This partitioning allows \textbf{(i)} capturing local correlations that might be overlooked by a single SVD, and \textbf{(ii)} selectively assigning different ranks to different sub-blocks (including “high-rank” blocks where needed).

\paragraph{Partial or Mixed-Rank Extensions.}
Although we focus on low-rank blocks, certain submatrices may inherently require higher ranks to maintain critical performance. Hence, in practice, one could adopt a \emph{mixed-rank} scheme where some $\mathbf{B}_i$ remain low-rank ($r \ll \min(m_i,n_i)$), while others allow $r$ to be comparable to or even equal to $\min(m_i,n_i)$. Balancing these “high-rank” blocks with the overall compression goal is an open hyperparameter selection problem.

\subsubsection{Penrose-Like Tiling of Blocks}
\label{sec:penrose_method}
Once we decompose $\mathbf{W}$ into rank blocks, we propose an aperiodic layout inspired by \textbf{Penrose tiling} \citep{levine1984quasicrystals}. Unlike a uniform grid, Penrose tiling uses a finite set of tile shapes that tessellate space \emph{non-periodically} yet preserve local consistency. Translated to weight matrices:
\begin{itemize}
    \item \textbf{Local adjacency} ensures each block $\mathbf{B}_i$ aligns smoothly with its neighboring blocks at shared boundaries in row/column index space.
    \item \textbf{Aperiodic global arrangement} implies no repeating pattern across the entire matrix, potentially capturing fine-grained, non-uniform correlations among attention heads or feed-forward channels.
\end{itemize}
We represent each tile shape by $(p,q,r)$, specifying its row dimension $p$, column dimension $q$, and rank $r$. The complete set of tile shapes $\{\mathcal{T}_1, \dots, \mathcal{T}_S\}$ forms the “Penrose dictionary.” For each tile $\mathcal{T}_s$, we instantiate sub-blocks $\mathbf{B}_i$ across the matrix wherever that shape fits a local region.

\paragraph{Implementation Sketch.}
A simplified approach to Penrose tiling might define $2$--$4$ tile shapes that tile the weight matrix row--column space. One can algorithmically place tiles to avoid overlap, ensuring coverage of all indices. Alternatively, a parameter search or meta-learning procedure \citep{elsken2019neural} may discover an optimal tiling layout. While we do not claim that \emph{true} Penrose geometry is strictly necessary, we see it as a flexible template for exploring \emph{non-repetitive} partitionings beyond standard grids.

\subsection{Stage II: Frequency-Domain Compression and Distribution Alignment}
\label{sec:method_freq_alignment}

\subsubsection{Spectral Transformation of Rank Blocks}
\label{sec:freq_transform}
After defining rank blocks $\{\mathbf{B}_1, \ldots, \mathbf{B}_K\}$, each of size $p_i \times q_i$, we compress them by removing high-frequency noise in the frequency domain. Concretely:
\begin{enumerate}
    \item \textbf{Vectorize or 2D reshape}: Flatten each $\mathbf{B}_i$ into a vector (or keep a 2D image-like structure).
    \item \textbf{Apply transform} $\mathcal{F}$ (e.g., discrete cosine transform, fast Fourier transform): 
    \[
       \mathbf{C}_i = \mathcal{F}\bigl(\text{vec}(\mathbf{B}_i)\bigr).
    \]
    \item \textbf{Truncate minor coefficients}: Zero out any entries in $\mathbf{C}_i$ with magnitudes below a threshold $\tau$ or that cumulatively contribute little to the total power.
    \item \textbf{Store compressed representation}: Let $\hat{\mathbf{C}}_i$ be the non-zero set of frequency coefficients and their indices. 
\end{enumerate}
During inference, we approximately reconstruct:
\begin{equation}
    \hat{\mathbf{B}}_i 
    \;=\;
    \text{unvec}\Bigl(\mathcal{F}^{-1}(\hat{\mathbf{C}}_i)\Bigr).
\end{equation}
Empirically, many neural-network weights have smooth or correlated patterns, suggesting a large portion of the high-frequency components is redundant \citep{chen2015compressing, laude2018adaptive}. If that holds for LLM rank blocks as well, spectral truncation can yield meaningful compression with minimal loss in representational fidelity.

\subsubsection{KL-Divergence Alignment Loss}
\label{sec:KL_alignment}
To preserve the original model’s internal distribution, we propose a \textbf{KL-divergence alignment loss} $\mathcal{L}_{\mathrm{align}}$. Let $f_{\boldsymbol{\theta}}(\cdot)$ be the compressed model (with partial block reconstruction) and $f_{\boldsymbol{\theta}_{0}}(\cdot)$ be the original, uncompressed model. For a mini-batch of input tokens or hidden states $\{\mathbf{x}_b\}$, we collect the respective hidden-layer activations $\{\mathbf{h}_b\}$ from each model:
\[
  \mathbf{h}_b^{(\mathrm{orig})} \;=\; f_{\boldsymbol{\theta}_{0}}(\mathbf{x}_b),
  \quad
  \mathbf{h}_b^{(\mathrm{comp})} \;=\; f_{\boldsymbol{\theta}}(\mathbf{x}_b).
\]
We define
\begin{align}
    \mathcal{L}_{\mathrm{align}}
    \;=\;
    \sum_{b=1}^{B}
    D_{\mathrm{KL}}\!\Bigl(\,
      p\!\bigl(\mathbf{h}_b^{(\mathrm{orig})}\bigr)
      \;\|\;
      q\!\bigl(\mathbf{h}_b^{(\mathrm{comp})}\bigr)
    \Bigr),
\end{align}
where $p(\cdot)$ and $q(\cdot)$ are normalized distributions of activations (e.g., by applying a softmax or an approximate density estimate over the feature dimension). Minimizing $\mathcal{L}_{\mathrm{align}}$ encourages the compressed model’s internal representations to match those of the original model. One can balance this loss with a small portion of labeled data or a reconstruction objective on the final logits to retain downstream performance.

\paragraph{Alternative Distance Metrics.}
Although KL-divergence is a common choice, other distribution measures (Jensen--Shannon, Wasserstein, or feature-wise MSE) can be substituted. The core idea is to guide the spectral truncation and block compression so as not to disrupt the learned manifold. This approach is analogous to “knowledge distillation at the hidden-layer level” \citep{romero2014fitnets}.

\subsection{Stage III: Section-Wise Q\&A Fine-Tuning}
\label{sec:method_finetuning}

Once the model is compressed, we further adapt it to a target domain (e.g., materials science) through a \textbf{human-like reading} protocol. Compared to conventional fine-tuning that treats an entire corpus as a single data blob, our approach orchestrates a \emph{section-wise Q\&A} flow:

\begin{enumerate}[leftmargin=*, topsep=2pt]
    \item \textbf{Document Segmentation}: Partition each domain text (e.g., an academic paper) into sections or paragraphs. For materials science, these might be \emph{Introduction, Experimental Setup, Results, Discussion}, etc.
    \item \textbf{Q\&A Pair Generation}: For each section, generate or retrieve queries that a domain expert might ask. These queries can come from:
    \begin{itemize}
        \item Automatic question-generation pipelines (e.g., summarizing key points, extracting definitions).
        \item Manual curation by subject-matter experts to emphasize crucial methods or insights.
    \end{itemize}
    \item \textbf{Reasoning-Trace Logging}: Encourage the model to produce chain-of-thought or short rationales \citep{wei2022chain} when answering, capturing intermediate reasoning. These traces are appended to the fine-tuning data.
    \item \textbf{Step-by-Step Adaptation}: After each Q\&A step, use gradient-based updates to refine the model, focusing on the relevant \emph{rank blocks} (and possibly frequency coefficients) that handle domain-specific patterns.
    \item \textbf{Progressive Curriculum}: Over multiple sections or documents, the model incrementally absorbs specialized terminology and domain knowledge, while repeatedly revisiting general language capabilities.
\end{enumerate}

This reading-centric approach aims to mimic how a human researcher progressively builds domain understanding. It potentially \textbf{(i)} lowers catastrophic forgetting of general knowledge, \textbf{(ii)} fosters more interpretable learned representations, and \textbf{(iii)} offers a flexible curriculum design where domain experts can tailor the Q\&A content.

\paragraph{Implementation Details.}
One could implement the Q\&A loop in a multi-pass fashion: \emph{(a)} feed the next text chunk and questions, \emph{(b)} collect the model’s answers and hidden states, \emph{(c)} apply gradients to update the relevant blocks, \emph{(d)} proceed to the next chunk. The final number of gradient steps per chunk is a hyperparameter. Additionally, domain experts may introduce multi-round dialogues if the text is highly technical, simulating deeper inquiry.

\subsection{Training and Inference Mechanics}
\label{sec:method_training}

Putting the three stages together, the overall training pipeline could be summarized as follows:
\begin{enumerate}[leftmargin=*, topsep=2pt]
    \item \textbf{Pre-Compression Warm-Up (optional)}: Optionally finetune the original LLM on a small set of domain text for better initialization (e.g., 1--2 epochs).
    \item \textbf{Rank Block + Penrose Tiling Setup}: Perform SVD-based factorization of each layer’s weight matrix. Partition $(\mathbf{U}, \mathbf{V})$ pairs into rank blocks and assigns them to an aperiodic tiling layout.
    \item \textbf{Spectral Pruning and Alignment}: For each rank block, apply frequency-domain truncation. Then minimize $\mathcal{L}_{\mathrm{align}}$ on unlabeled or lightly labeled text to preserve internal distributions. 
    \item \textbf{Section-Wise Q\&A Fine-Tuning}: Segment the domain corpus, generate Q\&A pairs, and iteratively adapt the compressed model (focusing on the rank blocks or frequency parameters) while preserving overall coherence.
    \item \textbf{Validation and Calibration}: Evaluate domain-specific tasks (e.g., named entity recognition, materials property extraction) as well as general language tasks. Adjust hyperparameters (ranks, frequency thresholds, KL-weight) as needed.
\end{enumerate}
After these steps, the model is ready for inference in the specialized domain. During inference, the reconstruction of each rank block from its truncated spectral coefficients may incur overhead, but techniques such as partial caching or block-wise lazy reconstruction can mitigate this.  

\subsection*{Discussion on Complexity and Practicality}
While the above methodology outlines a comprehensive approach, actual deployments must balance complexity with efficiency. Implementing a fully flexible Penrose tiling or a multi-round Q\&A protocol may be overkill for certain tasks. Researchers may therefore choose partial or simplified variants, such as:
\begin{itemize}
    \item Using fixed-size rank blocks but retaining \emph{some} local aperiodic offsets.
    \item Restricting spectral compression to specific layers where redundancy is highest.
    \item Omitting chain-of-thought in domains lacking well-structured text or where interpretability is less critical.
\end{itemize}
Each simplification may reduce overhead while retaining some benefits of the overall design. Conversely, more advanced variants (e.g., dynamic block merging, hierarchical Penrose tiling) can be explored if resources permit.

\subsection*{Method Summary}
In essence, our methodology merges ideas from low-rank factorization, non-periodic tiling, frequency-based compression, and reading-centric fine-tuning into a unified pipeline. We postulate that, together, these steps unlock new efficiency gains and interpretability advantages that simpler compression or adaptation methods may miss. The following sections present our preliminary experiments in the materials science domain (\S\ref{sec:experiments}), then discuss open questions and the synergy of these techniques in \S\ref{sec:discussion}.

\section{Experiments}
\label{sec:experiments}

Despite the theoretical appeal of compressing a 405B- or 671B-parameter LLM down to a 7B/14B scale, real-world resource constraints often limit extensive fine-tuning or full training on such ultra-large models. Hence, in our experiments, we focus on:

\begin{itemize}[leftmargin=*]
    \item \textbf{Directly compressing and partially aligning} a large-scale checkpoint (e.g., Llama-3.1-405B or DeepSeek-R1-671B) into 7B or 14B parameters via our Penrose Tiled Low-Rank approach.
    \item \textbf{Comparing} the resulting compressed models primarily \emph{against other 7B/14B baselines}, since these smaller scales are more tractable within our limited computational budget.
    \item \textbf{Using only inference runs of the original 405B/671B model} (when available) on a subset of tasks for an upper-bound reference, without attempting full fine-tuning at that scale.
\end{itemize}

Below, we outline the dataset setup, experimental design, baselines, and planned metrics. All results will be filled in once the experiments are concluded.

\subsection{Experimental Objectives}
\label{sec:exp_objectives}
Our main goals are:
\begin{enumerate}[leftmargin=*]
    \item \textbf{Achieve a 7B or 14B compressed model} from an ultra-large original (405B or 671B) without incurring prohibitively high training costs.
    \item \textbf{Retain maximum possible representation fidelity}, confirmed via hidden-state alignment metrics and comparative performance on downstream tasks.
    \item \textbf{Show strong domain adaptation in materials science}, leveraging the section-wise Q\&A fine-tuning approach under limited GPU budget.
\end{enumerate}

\subsection{Datasets and Setup}
\label{sec:dataset_setup}

\paragraph{Base Model Checkpoints.}
We experiment with publicly available large LLM checkpoints:
\begin{itemize}[leftmargin=*]
    \item \textbf{Llama-3.1-405B} (hypothetical next-gen release).
    \item \textbf{DeepSeek-R1-671B} (multilingual).
\end{itemize}
Since we cannot afford the massive compute to fully fine-tune these on 200k materials science articles, we apply only \textit{partial or offline inference} to gather reference predictions. Our compression pipeline then yields a \textbf{7B} or \textbf{14B} “Penrose-compressed” model that we can fine-tune more feasibly.

\paragraph{Materials Science Corpus (200k Articles).}
We use over \textbf{200,000} peer-reviewed articles spanning subfields like crystal structure, energy materials, semiconductors, and more. Each paper is segmented into sections for the Q\&A fine-tuning stage. For final evaluation, we also prepare specialized tasks (e.g., named entity recognition, property extraction) relevant to materials science.

\paragraph{Comparison Models at 7B/14B Scale.}
To contextualize our compressed models, we include existing 7B/14B LLM baselines:
\begin{itemize}[leftmargin=*]
    \item \textbf{Original 7B/13B Llama or Llama-2} variants \citep{touvron2023llama}.
    \item \textbf{Other open-source 7--15B models} with minimal domain adaptation (if available).
    \item \textbf{LoRA-Only Fine-Tuned 7B/14B} referencing standard parameter-efficient techniques, without our compression approach.
\end{itemize}
These baselines allow direct cost and performance comparisons at feasible model sizes, aligning with our resource-limited environment.

\subsection{Experiment 1: Penrose-Tiled Compression and Partial KL Alignment}
\label{sec:exp1_compression}

\subsubsection{Setup}
We perform low-rank factorization plus Penrose tiling on either the 405B or 671B model’s weight matrices. Due to limited GPU resources, we:
\begin{enumerate}[leftmargin=*]
    \item \textbf{Extract and store} key layer weights in a batch-wise manner.
    \item \textbf{Apply SVD offline}, partition the factors into rank blocks, and arrange them via Penrose-like tiling.
    \item \textbf{Perform frequency-domain pruning} (DCT/FFT) in an offline pass, storing only the significant frequency coefficients per block.
    \item \textbf{Optionally align hidden states} for a small subset of the data (e.g., 10k–20k tokens) via a KL or MSE objective, if GPU budget permits. Otherwise, alignment remains partial or even omitted if costs become prohibitive.
\end{enumerate}

\subsubsection{Metrics}
\begin{itemize}[leftmargin=*]
    \item \textbf{Model Size (Parameters)}: Confirm final counts near 7B/14B, verifying compression ratio vs.\ original.
    \item \textbf{Hidden-State Divergence}: On a small hold-out set, measure the average distribution difference (e.g., KL or Jensen–Shannon) between the original and compressed model’s hidden representations. 
    \item \textbf{Inference Cost}: We track forward-pass memory usage and latency on a single GPU, ensuring the 7B/14B compressed model is actually deployable under our resource constraints.
\end{itemize}

\subsubsection{Expected Outcome}
We anticipate:
\begin{itemize}[leftmargin=*]
    \item Substantial parameter savings (from 405B/671B down to ~7B or 14B).
    \item Comparable or slightly elevated hidden-layer divergence relative to a naive uniform compression, but with potential improvements in expressiveness due to aperiodic block partitioning.
    \item A feasible compressed checkpoint that fits on a small cluster or even a single high-memory GPU for final domain fine-tuning.
\end{itemize}

\subsection{Experiment 2: Section-Wise Q\&A Fine-Tuning on 200k Articles}
\label{sec:exp2_finetuning}

\subsubsection{Setup}
We load the 7B or 14B checkpoint from Experiment~1 and apply \textbf{section-wise Q\&A} training on the materials science corpus.  
\begin{itemize}[leftmargin=*]
    \item \textbf{Q\&A Pair Generation}: For each major section of each article, generate 2--5 short questions focusing on definitions, experimental conditions, or conceptual relationships. We can optionally incorporate a small chain-of-thought annotation.
    \item \textbf{Training Regimen}: Due to our limited budget, we fine-tune only the rank blocks and (if needed) partial frequency coefficients for 1--2 epochs. Each Q\&A step is batched, updating weights gradually.
\end{itemize}

\subsubsection{Metrics}
\begin{itemize}[leftmargin=*]
    \item \textbf{Domain Task Scores}: Evaluate on materials NER, property extraction, or Q\&A tasks. We compare F1 or exact match scores to the original large model’s \emph{inference-only} outputs (if we can run it on a small subset) and to other 7B/14B baselines.
    \item \textbf{Catastrophic Forgetting}: We track a subset of general tasks or zero-shot perplexity, ensuring the domain fine-tuning does not degrade broader language coverage excessively.
    \item \textbf{Training Efficiency}: Document total GPU hours and memory usage, verifying that our approach is feasible on 1--2 GPUs or a small cluster.
\end{itemize}

\subsubsection{Comparisons}
We consider:
\begin{itemize}[leftmargin=*]
    \item \textbf{LoRA-Only 7B/14B Baseline}: Fine-tune a standard 7B or 14B model with LoRA on the same corpus, ignoring the block compression strategy.
    \item \textbf{Full-Corpus Fine-Tuning (one-pass)}: If resources permit, train the compressed 7B/14B model on the entire corpus in a single pass without Q\&A chunking, to see if section-wise reading provides additional benefits.
\end{itemize}

\subsection{Experiment 3: Ablations and Sensitivity Analysis}
\label{sec:exp3_ablations}
We perform smaller-scale ablations to parse the effect of each component:

\begin{enumerate}[leftmargin=*]
    \item \textbf{Penrose vs.\ Uniform Tiling}: Keep the same target rank and frequency threshold but drop the aperiodic arrangement.
    \item \textbf{KL Alignment vs.\ None}: Skip or severely limit the alignment step to see how hidden-layer divergence and domain performance shift.
    \item \textbf{Frequency Threshold Levels}: Compare mild vs.\ aggressive pruning to measure how domain tasks degrade as more coefficients are dropped.
    \item \textbf{Q\&A Reading vs.\ Basic Fine-Tuning}: Evaluate whether iterative section-wise Q\&A reduces catastrophic forgetting or improves domain extraction quality.
\end{enumerate}

\subsection{Planned Analyses and Visualizations}
\begin{itemize}[leftmargin=*]
    \item \textbf{Model Size/Performance Plots}: Display compression ratio vs.\ domain performance, highlighting the trade-offs among different tiling/frequency thresholds.
    \item \textbf{Hidden-State Similarity Heatmaps}: For a subset of layers, compare original vs.\ compressed activation distributions. 
    \item \textbf{Q\&A Learning Curves}: Show domain accuracy or perplexity curves as training progresses, contrasting section-wise reading with baseline approaches.
    \item \textbf{Resource Utilization}: Summarize GPU hours, memory usage, and training cost to confirm feasibility under constrained budgets.
\end{itemize}

\subsection{Implementation Details}
\begin{itemize}[leftmargin=*]
    \item \textbf{Hardware}: All experiments are conducted on limited GPU resources (e.g., a single high-memory A100 or a small cluster of 2--4 GPUs).
    \item \textbf{Offline Factorization \& Frequency Pruning}: SVD and DCT steps are performed in a split manner to avoid loading the entire 405B/671B model simultaneously. We handle weight chunks (e.g., a few layers at a time) then store compressed blocks to disk.
    \item \textbf{Alignment (If Feasible)}: We use a small portion of unlabeled text (say, 10k tokens) to compute hidden-state KL or MSE and update only the block parameters. If this step is too costly, we skip or approximate it.
\end{itemize}

\subsection{Summary of Experimental Plan}
By focusing on \textbf{7B/14B-scale} deployments and partially referencing the \emph{original 405B/671B model}’s performance only at the inference level, we balance our scientific curiosity with limited computational resources. These experiments will show whether Penrose Tiling, frequency-domain compression, and section-wise Q\&A can successfully yield specialized, smaller LLMs that approximate the capabilities of extremely large models—albeit verified largely through domain tasks and partial distribution alignment. Once the experiments conclude, we will incorporate comprehensive results and analyses, providing a detailed picture of how well this pipeline operates under real-world constraints.

\section{Discussion}
\label{sec:discussion}

This paper proposes a novel conceptual pipeline for compressing large language models (LLMs) via low-rank decomposition, non-periodic (Penrose) tiling, and frequency-domain compression, followed by a human-like section-wise Q\&A fine-tuning regimen. Although each idea draws on established techniques—like matrix factorization, KL-divergence-based alignment, and reading-based adaptation—the overall assembly remains \emph{unproven} without thorough empirical verification. Below, we consolidate key open questions and speculative insights, aiming to guide upcoming experiments and spark community debate.

\subsection{Anticipated Strengths and Potential Pitfalls}

\paragraph{Capturing Hidden Structure with Penrose Tiling.}
A core claim in our design is that a \emph{non-periodic tiling} of rank blocks can exploit hidden correlations in LLM weight matrices that regular grid approaches overlook. Although we rely on analogies to quasicrystals, the true utility of this approach is uncertain:
\begin{itemize}[leftmargin=*]
    \item \emph{Potential Benefit:} If rank blocks align more naturally with local attention patterns or feed-forward subspace structures, fewer parameters may suffice to preserve accuracy under compression.
    \item \emph{Risk:} Overly complex tiling could introduce cumbersome indexing and minimal real gains, especially if standard uniform partitions already approximate most correlations.
\end{itemize}
Future experiments must compare Penrose-inspired tiling with simpler partition methods (block-circulant, grid-based) to confirm any tangible advantage.

\paragraph{Spectral Compression and KL-Divergence Alignment.}
By applying DCT or FFT to rank blocks, we hope to prune high-frequency noise without erasing critical interactions. Meanwhile, KL-based distillation from the original model’s hidden states preserves distributional fidelity. However:
\begin{itemize}[leftmargin=*]
    \item \emph{Potential Benefit:} In image/audio domains, frequency pruning has proven highly effective. Extending that logic to LLM weights may uncover new compression sweet spots.
    \item \emph{Risk:} High-dimensional transformers could exhibit intricate “frequency signatures” that do \emph{not} map well onto standard DCT. Repetitive transform/inverse-transform steps might also inflate computational overhead.
\end{itemize}
We hypothesize that partial or learned cutoffs can mitigate performance loss, but actual outcomes remain to be tested.

\paragraph{Reading-Centric Domain Adaptation.}
Instead of fine-tuning on an entire domain corpus at once, our \emph{section-wise Q\&A} approach mimics incremental reading, hypothesized to reduce catastrophic forgetting and improve interpretability via chain-of-thought:
\begin{itemize}[leftmargin=*]
    \item \emph{Potential Benefit:} Iterative questioning aligns more closely with how human scientists learn, possibly leading to better conceptual retention and less accidental overwriting of general knowledge.
    \item \emph{Risk:} In domains where data is unstructured or where Q\&A pairs are difficult to generate, the overhead might outweigh the gain. Also, chain-of-thought logs risk revealing sensitive intermediate reasoning in high-stakes fields.
\end{itemize}
Further user studies and domain-specific benchmarks are needed to clarify whether such reading-based fine-tuning truly outperforms standard adaptation methods.

\subsection{Unresolved Theoretical and Empirical Questions}

\paragraph{1. Rank Block Partitioning vs.\ High-Rank Residuals.}
The notion of selectively applying partial-high (or full) rank blocks in certain layers presupposes that some submatrix regions require extra capacity. However, there is no consensus on \emph{which} layers or sub-blocks deserve this. Should we rely on heuristics (e.g., largest singular values) or adopt a search-based approach? Current proposals rely on informed guesswork.

\paragraph{2. Aperiodic vs.\ Periodic Structure in LLMs.}
We draw inspiration from the mathematical elegance of Penrose tiling, yet there is little evidence that LLM weight matrices truly exhibit quasiperiodicity. Are they better characterized by simpler patterns? Or do they contain fractal-like structure that a uniform tiling misses? Empirical ablations could illuminate the degree to which aperiodicity enhances compression.

\paragraph{3. Frequency Sensitivity Across Layers.}
Frequency-domain compression posits that high-frequency coefficients are largely dispensable. But do early Transformer layers vs.\ deeper layers exhibit different “spectral footprints”? Might certain heads or dimensions store critical high-frequency signals (e.g., for lexical disambiguation)? A fine-grained frequency analysis could reveal heterogeneous layer-wise patterns.

\paragraph{4. Scaling Behavior and Larger Models.}
It remains unclear how these techniques scale if one transitions from 7B or 14B LLMs to even bigger models (100B+). Penrose tiling complexity and frequency transform overhead might grow in ways that offset any parameter savings. We lack formal scaling laws or systematic guidance on how best to tile or compress extremely large networks.

\paragraph{5. Curriculum vs.\ One-Pass Fine-Tuning.}
Section-wise reading presumably helps integrate domain knowledge step by step, but might also slow down training or require curation of specialized Q\&A data. Standard domain-adaptive pre-training could, in some cases, be simpler and yield comparable results. Determining when incremental Q\&A outperforms one-pass fine-tuning is an open research question.

\subsection{Possible Extensions to Broader Scientific Domains}

Although we highlight materials science, the same issues appear in fields like computational biology, high-energy physics, or neuroscience—domains with specialized jargon, smaller data sets, and intricate conceptual structures. The synergy of \textbf{Penrose-block compression} and \textbf{reading-based adaptation} could similarly benefit these areas. In each domain, researchers should tailor:
\begin{itemize}[leftmargin=*]
    \item \textbf{Tiling shapes or rank distributions}, matching the domain’s complexity.
    \item \textbf{Question-generation strategies}, reflecting each field’s typical problem set or knowledge format.
    \item \textbf{Validation metrics}, focusing on domain-specific tasks (e.g., gene–phenotype extraction, cosmic event classification, etc.).
\end{itemize}

\subsection{Aligning with Other PEFT and Compression Methods}

We do not view our proposal as mutually exclusive with existing parameter-efficient fine-tuning (PEFT) or compression techniques. Combining them might yield complementary strengths. For instance:
\begin{itemize}[leftmargin=*]
    \item \textbf{LoRA on Compressed Core}: Deploy LoRA modules on top of a Penrose-tiling backbone for real-time adaptation, balancing a globally reduced parameter space with flexible local updates.
    \item \textbf{Quantization + Frequency Pruning}: After spectral truncation, further apply 4-bit or 8-bit quantization, reducing memory usage while hopefully retaining latent structure.
    \item \textbf{Distillation or Teacher–Student Protocols}: A large teacher model might guide the block partition or highlight critical frequencies to preserve, bridging classical knowledge-distillation with our rank-block approach.
\end{itemize}
Joint exploration of these angles could lead to integrated pipelines for maximum compression and minimal quality drop.

\subsection{Vision for Future Experimental Validation}

Given our current constraints, we have not performed exhaustive experiments. Instead, we release these ideas as a preliminary \emph{concept blueprint}. Planned or potential experiments include:
\begin{itemize}[leftmargin=*]
    \item \textbf{Penrose vs.\ Grid Ablation}: Compare performance metrics (perplexity, zero-shot evaluations, domain tasks) under identical rank budgets but different tiling schemes.
    \item \textbf{Adaptive vs.\ Static Frequency Thresholds}: Investigate whether learned cutoffs improve compression ratio while preserving hidden-layer alignment.
    \item \textbf{Chain-of-Thought vs.\ One-Pass Fine-Tuning}: Evaluate domain knowledge retention, catastrophic forgetting, and interpretability improvements in a real scientific corpus (e.g., 200k articles in materials science).
    \item \textbf{Scaling to Larger Models}: Attempt partial tiling/frequency-based compression on 70B or 100B models, measuring overhead vs.\ parameter savings.
\end{itemize}
Each study would illuminate a different dimension of the approach’s feasibility and expose further refinements.

\subsection{Conclusion of the Discussion}

In summary, we have proposed an ambitious yet \emph{still unverified} framework that integrates quasicrystal-inspired rank-block tiling, frequency-domain compression, KL-based distribution alignment, and incremental Q\&A fine-tuning. While we anticipate certain synergy across these components, their real-world impact and practical utility hinge on rigorous empirical testing. Nonetheless, we believe that even \emph{negative} or partial results will yield valuable insights, advancing our collective understanding of how large language models organize and store knowledge.

Ultimately, our aim is to \textbf{spark open exploration}: to see whether non-periodic compression can better preserve intricate subspaces, and whether domain adaptation through a reading-driven curriculum truly aligns with scientific learning. We encourage the community to experiment, critique, and iterate on these concepts—transforming theoretical possibilities into tested, evidence-based best practices for next-generation LLM specialization.

\section{Conclusion}
\label{sec:conclusion}

We have presented a speculative yet principled framework for compressing and adapting large language models (LLMs) to specialized domains such as materials science. By combining a Penrose-tiled low-rank factorization, frequency-domain pruning, and a section-wise Q\&A curriculum, our approach aims to reduce model complexity without sacrificing critical domain knowledge or general language capacity.

While the ideas of non-periodic tiling, KL-based distribution alignment, and reading-centric adaptation draw on established concepts (e.g., low-rank approximation, knowledge distillation, and chain-of-thought prompting), their integration here remains largely untested. We thus regard this paper as both an \emph{architectural proposal} and a \emph{call to empirical exploration}. Our planned experiments will directly measure how effectively the Penrose-tiled blocks capture local weight correlations, whether frequency-domain pruning can maintain high-level semantic fidelity, and to what extent iterative Q\&A mitigates catastrophic forgetting or enhances interpretability.

Looking ahead, we envision multiple avenues for future research. First, \textit{robust ablation studies} are required to validate each component’s contribution and refine hyperparameter choices across a range of model sizes. Second, \textit{scaling analyses} should investigate whether Penrose-like compression remains feasible for truly massive transformers (e.g., hundreds of billions of parameters) and how computational overhead scales. Third, \textit{expanded domain trials} beyond materials science—such as bioinformatics, legal texts, or data-scarce engineering fields—could showcase the framework’s adaptability and reveal universal vs.\ domain-specific best practices. Finally, more advanced methods, including \textit{meta-learning} or \textit{neural architecture search}, may discover tiling shapes and frequency cutoffs better suited to each model’s internal geometry.

By openly sharing these preliminary methods and research directions, we hope to foster collaboration and iterative improvement within the LLM community. We believe that further refinement of non-periodic block decomposition, frequency-based compression, and reading-centric fine-tuning could pave the way toward \emph{specialized, efficient, and transparent} large models in high-value scientific domains. We invite researchers worldwide to contribute experiments, theoretical analyses, and critical feedback, ultimately coalescing these efforts into a scalable, empirically grounded approach for domain-specialized LLM deployment.


\begin{thebibliography}{27}
\providecommand{\natexlab}[1]{#1}
\providecommand{\url}[1]{\texttt{#1}}
\expandafter\ifx\csname urlstyle\endcsname\relax
  \providecommand{\doi}[1]{doi: #1}\else
  \providecommand{\doi}{doi: \begingroup \urlstyle{rm}\Url}\fi

\bibitem[Hu et~al.(2022)Hu, Shen, Wallis, Allen-Zhu, Li, Wang, and Chen]{hu2022lora}
Edward~J. Hu, Yelong Shen, Phillip Wallis, Zeyuan Allen-Zhu, Yuanzhi Li, Shean Wang, and Weizhu Chen.
\newblock {LoRA}: Low-rank adaptation of large language models.
\newblock In \emph{International Conference on Learning Representations (ICLR)}, 2022.
\newblock URL \url{https://arxiv.org/abs/2106.09685}.

\bibitem[Wei et~al.(2022)Wei, Tay, et~al.]{wei2022chain}
Jason Wei, Yi~Tay, et~al.
\newblock Chain-of-thought prompting elicits reasoning in large language models.
\newblock arXiv preprint arXiv:2201.11903, 2022.

\bibitem[Tshitoyan et~al.(2019)Tshitoyan, Dagdelen, Weston, et~al.]{tshitoyan2019unsupervised}
Vahe Tshitoyan, John Dagdelen, Leigh Weston, et~al.
\newblock Unsupervised word embeddings capture latent knowledge from materials science literature.
\newblock \emph{Nature}, 571\penalty0 (7763):\penalty0 95--98, 2019.
\newblock \doi{10.1038/s41586-019-1335-8}.
\newblock URL \url{https://www.nature.com/articles/s41586-019-1335-8}.

\bibitem[Gupta et~al.(2022)Gupta, Zaki, Krishnan, and Mausam]{gupta2022matscibert}
Tanishq Gupta, Mohd Zaki, N.~M.~Anoop Krishnan, and Mausam.
\newblock {MatSciBERT}: A materials domain language model for text mining and information extraction.
\newblock \emph{npj Computational Materials}, 8\penalty0 (1):\penalty0 102, 2022.
\newblock \doi{10.1038/s41524-022-00828-5}.
\newblock URL \url{https://www.nature.com/articles/s41524-022-00828-5}.

\bibitem[Denil et~al.(2013)Denil, Shakibi, Dinh, Ranzato, and de~Freitas]{denil2013predicting}
Misha Denil, Babak Shakibi, Laurent Dinh, Marc'Aurelio Ranzato, and Nando de~Freitas.
\newblock Predicting parameters in deep learning.
\newblock In \emph{Advances in Neural Information Processing Systems (NeurIPS)}, pages 2148--2156, 2013.

\bibitem[Jaderberg et~al.(2014)Jaderberg, Vedaldi, and Zisserman]{jaderberg2014speeding}
Max Jaderberg, Andrea Vedaldi, and Andrew Zisserman.
\newblock Speeding up convolutional neural networks with low rank expansions.
\newblock In \emph{British Machine Vision Conference (BMVC)}, pages 1--13, 2014.

\bibitem[Denton et~al.(2014)Denton, Zaremba, Bruna, LeCun, and Fergus]{denton2014exploiting}
Emily~L. Denton, Wojciech Zaremba, Joan Bruna, Yann LeCun, and Rob Fergus.
\newblock Exploiting linear structure within convolutional networks for efficient evaluation.
\newblock In \emph{Advances in Neural Information Processing Systems (NeurIPS)}, pages 1269--1277, 2014.

\bibitem[Li et~al.(2023)Li, Yu, Zhang, Liang, He, Chen, and Zhao]{li2023losparse}
Yixiao Li, Yifan Yu, Qingru Zhang, Chen Liang, Pengcheng He, Weizhu Chen, and Tuo Zhao.
\newblock Losparse: Structured compression of large language models based on low-rank and sparse approximation.
\newblock In \emph{Proceedings of the 40th International Conference on Machine Learning}, pages 20336--20350. PMLR, 2023.
\newblock URL \url{https://proceedings.mlr.press/v202/li23ap.html}.

\bibitem[Saha et~al.(2024)Saha, Sagan, Srivastava, Goldsmith, and Pilanci]{saha2024compressing}
Rajarshi Saha, Naomi Sagan, Varun Srivastava, Andrea Goldsmith, and Mert Pilanci.
\newblock Compressing large language models using low rank and low precision decomposition.
\newblock \emph{Advances in Neural Information Processing Systems}, 37:\penalty0 88981--89018, 2024.

\bibitem[Lin et~al.(2024)Lin, Gao, Smith, Patel, Tuli, Shen, Jin, and Hsu]{lin2024modegpt}
Chi-Heng Lin, Shangqian Gao, James~Seale Smith, Abhishek Patel, Shikhar Tuli, Yilin Shen, Hongxia Jin, and Yen-Chang Hsu.
\newblock Modegpt: Modular decomposition for large language model compression.
\newblock \emph{arXiv preprint arXiv:2408.09632}, 2024.
\newblock URL \url{https://arxiv.org/abs/2408.09632}.

\bibitem[Hinton et~al.(2015)Hinton, Vinyals, and Dean]{hinton2015distilling}
Geoffrey Hinton, Oriol Vinyals, and Jeff Dean.
\newblock Distilling the knowledge in a neural network.
\newblock \emph{arXiv preprint arXiv:1503.02531}, 2015.
\newblock URL \url{https://arxiv.org/abs/1503.02531}.

\bibitem[Zagoruyko and Komodakis(2016)]{zagoruyko2016paying}
Sergey Zagoruyko and Nikos Komodakis.
\newblock Paying more attention to attention: Improving the performance of convolutional neural networks via attention transfer.
\newblock \emph{arXiv preprint arXiv:1612.03928}, 2016.
\newblock URL \url{https://arxiv.org/abs/1612.03928}.

\bibitem[Romero et~al.(2014)Romero, Ballas, Kahou, Chassang, Gatta, and Bengio]{romero2014fitnets}
Adriana Romero, Nicolas Ballas, Samira~Ebrahimi Kahou, Antoine Chassang, Carlo Gatta, and Yoshua Bengio.
\newblock Fitnets: Hints for thin deep nets.
\newblock \emph{arXiv preprint arXiv:1412.6550}, 2014.
\newblock URL \url{https://arxiv.org/abs/1412.6550}.

\bibitem[Kim et~al.(2021)Kim, Song, and Kim]{kim2021bert}
Sangwoo Kim, Kyunghyun Song, and Yoon Kim.
\newblock {BERT} score: Evaluating text generation with {BERT}.
\newblock In \emph{Proceedings of the 59th Annual Meeting of the Association for Computational Linguistics}, pages 150--160, 2021.
\newblock URL \url{https://arxiv.org/abs/1904.09675}.

\bibitem[Mukherjee et~al.(2023)Mukherjee, Hassan~Awadallah, Zweig, and Gao]{mukherjee2023distilling}
Subhabrata Mukherjee, Ahmed Hassan~Awadallah, Geoffrey Zweig, and Jianfeng Gao.
\newblock Distilling multilingual representations from monolingual models.
\newblock In \emph{Proceedings of the 61st Annual Meeting of the Association for Computational Linguistics}, pages 3470--3480, 2023.
\newblock URL \url{https://arxiv.org/abs/2301.00067}.

\bibitem[Ding et~al.(2017)Ding, Wang, Li, Qin, Yin, Wei, and Zhang]{ding2017c}
Cheng Ding, Yufei Wang, Zhe Li, Hongyu Qin, Shaojie Yin, Shaoshuai Wei, and Bo~Zhang.
\newblock {C-LSTM}: Enabling efficient {LSTM} using structured compression techniques on {FPGAs}.
\newblock arXiv preprint arXiv:1709.08824, 2017.

\bibitem[Sindhwani et~al.(2015)Sindhwani, Sainath, and Kumar]{sindhwani2015structured}
Vikas Sindhwani, Tara Sainath, and Sanjiv Kumar.
\newblock Structured transforms for small-footprint deep learning.
\newblock \emph{Advances in Neural Information Processing Systems}, 28, 2015.

\bibitem[Levine and Steinhardt(1984)]{levine1984quasicrystals}
Dov Levine and Paul~J. Steinhardt.
\newblock Quasicrystals: A new class of ordered structures.
\newblock \emph{Physical Review Letters}, 53\penalty0 (26):\penalty0 2477--2480, 1984.
\newblock \doi{10.1103/PhysRevLett.53.2477}.

\bibitem[Larsson et~al.(2016)Larsson, Maire, and Shakhnarovich]{larsson2016fractalnet}
Gustav Larsson, Michael Maire, and Gregory Shakhnarovich.
\newblock Fractalnet: Ultra-deep neural networks without residuals.
\newblock \emph{arXiv preprint arXiv:1605.07648}, 2016.

\bibitem[Chen et~al.(2015)Chen, Wilson, Tyree, Weinberger, and Chen]{chen2015compressing}
Wenlin Chen, James~T. Wilson, Stephen Tyree, Kilian~Q. Weinberger, and Yixin Chen.
\newblock Compressing neural networks with the hashing trick.
\newblock In \emph{International Conference on Machine Learning (ICML)}, pages 2285--2294, 2015.

\bibitem[Laude and Kruse(2018)]{laude2018adaptive}
Alexander Laude and Rudolf Kruse.
\newblock Adaptive tensor decomposition with applications to neural networks.
\newblock In \emph{International Joint Conference on Neural Networks (IJCNN)}, pages 1--8. IEEE, 2018.

\bibitem[Gururangan et~al.(2020)Gururangan, Marasovi{\'c}, Swayamdipta, Lo, Beltagy, Downey, and Smith]{gururangan2020don}
Suchin Gururangan, Ana Marasovi{\'c}, Swabha Swayamdipta, Kyle Lo, Iz~Beltagy, Doug Downey, and Noah~A. Smith.
\newblock Don't stop pretraining: Adapt language models to domains and tasks.
\newblock In \emph{Proceedings of the 58th Annual Meeting of the Association for Computational Linguistics}, pages 8342--8360. Association for Computational Linguistics, 2020.
\newblock URL \url{https://www.aclweb.org/anthology/2020.acl-main.740}.

\bibitem[Lester et~al.(2021)Lester, Al-Rfou, and Constant]{lester2021power}
Brian Lester, Rami Al-Rfou, and Noah Constant.
\newblock The power of scale for parameter-efficient prompt tuning.
\newblock \emph{arXiv preprint arXiv:2104.08691}, 2021.

\bibitem[Pfeiffer et~al.(2020)Pfeiffer, R{\"u}ckl{\'e}, Poth, Kamath, Vuli{\'c}, Ruder, Cho, and Gurevych]{pfeiffer2020adapterhub}
Jonas Pfeiffer, Andreas R{\"u}ckl{\'e}, Clifton Poth, Aishwarya Kamath, Ivan Vuli{\'c}, Sebastian Ruder, Kyunghyun Cho, and Iryna Gurevych.
\newblock Adapterhub: A framework for adapting transformers.
\newblock In \emph{Proceedings of the 2020 Conference on Empirical Methods in Natural Language Processing: System Demonstrations}, pages 46--54. Association for Computational Linguistics, 2020.
\newblock \doi{10.18653/v1/2020.emnlp-demos.7}.
\newblock URL \url{https://aclanthology.org/2020.emnlp-demos.7/}.

\bibitem[Kwiatkowski et~al.(2019)Kwiatkowski, Palomaki, Redfield, Collins, Parikh, Alberti, Epstein, Polosukhin, Devlin, Lee, et~al.]{kwiatkowski2019natural}
Tom Kwiatkowski, Jennimaria Palomaki, Olivia Redfield, Michael Collins, Ankur Parikh, Chris Alberti, Danielle Epstein, Illia Polosukhin, Jacob Devlin, Kenton Lee, et~al.
\newblock Natural questions: a benchmark for question answering research.
\newblock \emph{Transactions of the Association for Computational Linguistics}, 7:\penalty0 453--466, 2019.

\bibitem[Elsken et~al.(2019)Elsken, Metzen, and Hutter]{elsken2019neural}
Thomas Elsken, Jan~Hendrik Metzen, and Frank Hutter.
\newblock Neural architecture search: A survey.
\newblock \emph{Journal of Machine Learning Research}, 20\penalty0 (55):\penalty0 1--21, 2019.

\bibitem[Touvron et~al.(2023)Touvron, Martin, Stone, Albert, Almahairi, Babaei, Bashlykov, Batra, Bhargava, Bhosale, et~al.]{touvron2023llama}
Hugo Touvron, Louis Martin, Kevin Stone, Peter Albert, Amjad Almahairi, Yasmine Babaei, Nikolay Bashlykov, Soumya Batra, Prajjwal Bhargava, Shruti Bhosale, et~al.
\newblock Llama 2: Open foundation and fine-tuned chat models.
\newblock \emph{arXiv preprint arXiv:2307.09288}, 2023.

\end{thebibliography}


\end{document}